\documentclass[lettersize,journal]{IEEEtran}
\usepackage{amsmath,amsfonts}
\usepackage{algorithmic}
\usepackage{algorithm}
\usepackage{array}
\usepackage[caption=false,font=normalsize,labelfont=sf,textfont=sf]{subfig}
\usepackage{textcomp}
\usepackage{stfloats}
\usepackage{url}
\usepackage{verbatim}
\usepackage{graphicx}
\usepackage{natbib}
\usepackage{hyperref}
\usepackage{xcolor}
\usepackage{multirow}
\usepackage{threeparttable}
\emergencystretch=\maxdimen
\setcitestyle{numbers,square}
\title{EEPNet-V2: Patch-to-Pixel Solution for Efficient Cross-Modal Registration between LiDAR Point Cloud and Camera Image}
\author{Yuanchao Yue, Hui Yuan, Senior Member, IEEE, Zhengxin Li, Shuai Li, Senior Member, IEEE, Wei Zhang, Senior Member, IEEE
\thanks{This work was supported in part by the National Natural Science Foundation of China under Grants 62222110 and 62172259,  the High-end Foreign Experts Recruitment Plan of Chinese Ministry of Science and Technology under Grant G2023150003L, the Taishan Scholar Project of Shandong Province (tsqn202103001), the Natural Science Foundation of Shandong Province under Grant ZR2022ZD38, Key Technology Research and Development Program of Shandong Province 2024CXGC010212.(\textit{Corresponding author: Hui Yuan.})

Yuanchao Yue, Hui Yuan, Zhengxin Li, Shuai Li and Wei Zhang are with the School of Control Science and Engineering, Shandong University, Jinan, Shandong, China, and also with the Key Laboratory of Machine Intelligence and System Control, Ministry of Education, Jinan 250061, China (e-mail: huiyuan@sdu.edu.cn).

}}
\date{May 2024}
\begin{document}

\maketitle

\begin{abstract}
The primary requirement for cross-modal data fusion is the precise alignment of data from different sensors. However, the calibration between LiDAR point clouds and camera images is typically time-consuming and needs external calibration board or specific environmental features. Cross-modal registration effectively solves this problem by aligning the data directly without requiring external calibration. However, due to the domain gap between the point cloud and the image, existing methods rarely achieve satisfactory registration accuracy while maintaining real-time performance. To address this issue, we propose a framework that projects point clouds into several 2D representations for matching with camera images, which not only leverages the geometric characteristic of LiDAR point clouds effectively but also bridge the domain gap between the point cloud and image. Moreover, to tackle the challenges of cross modal differences and the limited overlap between LiDAR point clouds and images in the image matching task, we introduce a multi-scale feature extraction network to effectively extract features from both camera images and the projection maps of LiDAR point cloud. Additionally, we propose a patch-to-pixel matching network to provide more effective supervision and achieve high accuracy. We validate the performance of our model through experiments on the KITTI and nuScenes datasets. Experimental results demonstrate the the proposed method achieves real-time performance and extremely high registration accuracy. Specifically, on the KITTI dataset, our model achieves a registration accuracy rate of over 99\%. Our code is released at: https://github.com/ESRSchao/EEPNet-V2.
\end{abstract}


\section{Introduction}
In autonomous driving and robot systems, the fusion between data collected from different sensors now plays an important role in many downstream tasks, such as object detection \cite{bai2022transfusion, chen2017multi, ku2018joint} and segmentation \cite{caltagirone2021lidar, wu2018squeezeseg, wang2018depth}. Better accuracy and efficiency can be achieved by using these fused data, thereby ensuring safety and robustness in complex environments. Among the data collected from different sensors, the most common process is to perform a fusion of the LiDAR (solid-state LiDAR or rotating LiDAR, etc.) point cloud and the camera (high-definition cameras or infrared cameras, etc.) image. 
\begin{figure*}
\centering
    \includegraphics[width=0.95\linewidth]{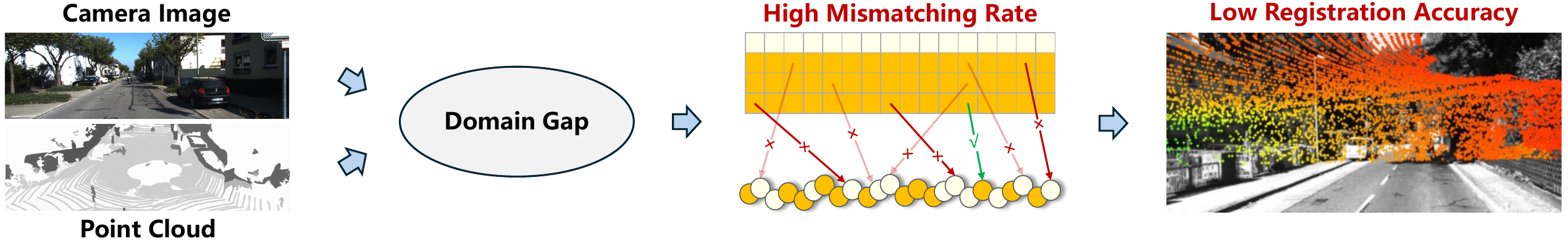} 
    \caption{The domain gap in the cross-modal registration task. Due to the dimensional differences between point clouds and images, the matching failure rate is high, which in turn reduces the registration accuracy.  The final graph is the projection result using a wrongly estimated extrinsic matrix}
    \label{fig:1}
\end{figure*}

However, the primary condition of a successful fusion is an accurate alignment between these data. A commonly used method to deal with the alignment task between sensors is to calculate the external parameters between sensors using a checkerboard \cite{kim2019extrinsic}, which we call off-line calibration. High-accuracy alignment between sensors by using these types of methods and unavoidable time costs. In addition, some online calibration methods \cite{schneider2017regnet, iyer2018calibnet, yuan2020rggnet} are proposed to correct external parameter errors caused by bumps or vibrations. These online calibration methods can modify the deviations within a limited error range. However, they will lose effectiveness when the errors are too large. Although these methods can achieve high-accuracy alignment, these complicated calibration processes make data fusion preprocessing time-consuming and inefficient. Whenever the deployment of sensors has changed, these calibration processes have to be repeated. 

To save manpower and resources, a straightforward way is to bypass these calibration processes by directly registering the data collected data from different sensors. Therefore cross-modal registration methods have emerged to address this challenge. Recent cross-modal registration methods \cite{feng20192d3d, pham2020lcd} can solve this alignment task by using specific deep learning networks. However, the cross-modal registration process typically involves comparing the similarity of the high-dimensional features between point clouds and images to establish correspondences, and the domain gap between 3D and 2D data always influence this process, as shown in Fig.  \ref{fig:1}. Specifically, there is a dimensional and informational difference between the 3D spatial data of the point cloud and the 2D texture information of the image. Although some existing methods attempt to decline these influence by exchanging feature information \cite{ren2022corri2p} or using feature extraction networks with more similar structures \cite{zhou2024differentiable}, this domain gap still affect the accuracy of the registration, especially leading to a high mismatching rate.

\begin{figure}
\centering
    \includegraphics[width=1\linewidth]{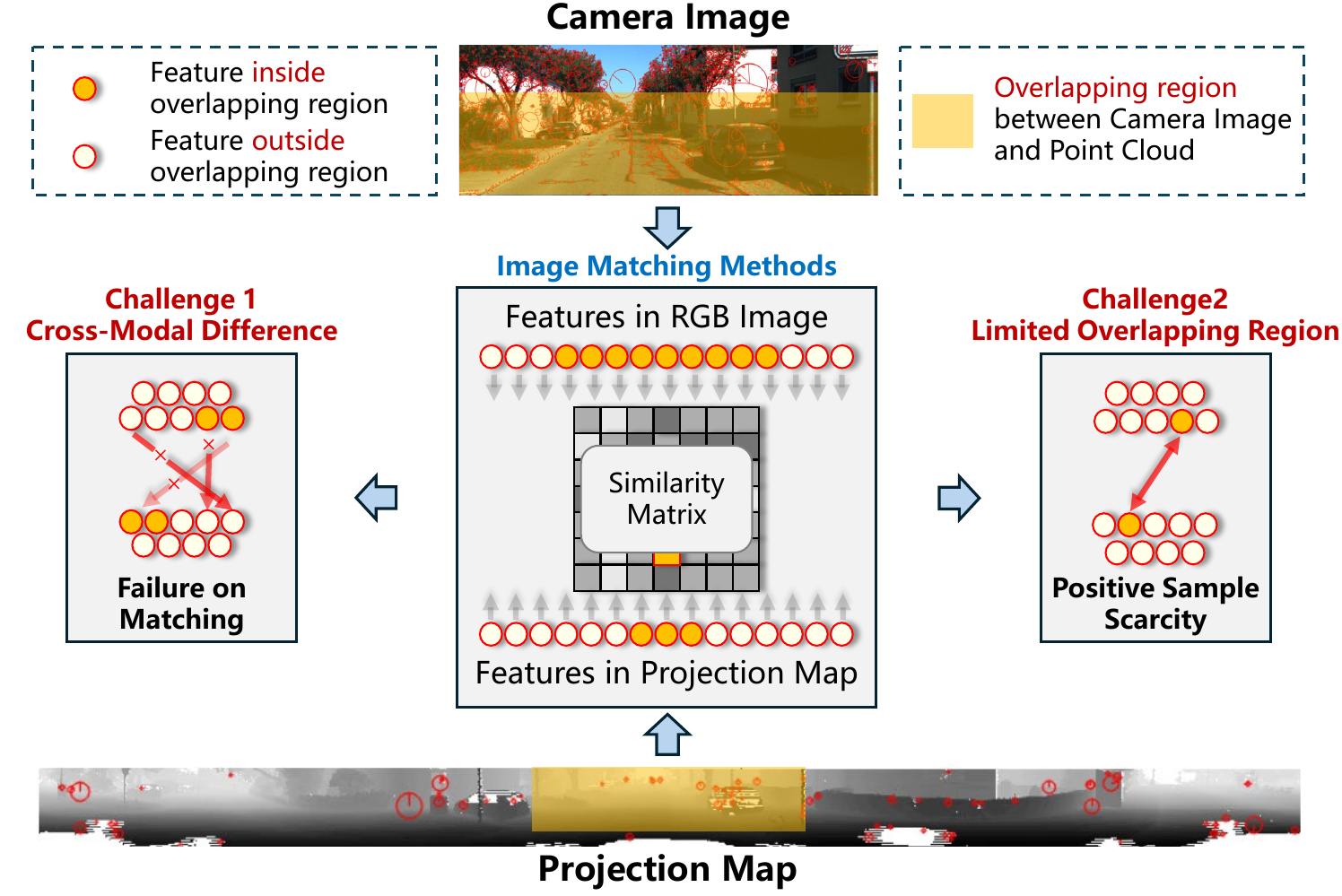}
    \caption{The challenge we faced in the matching task between projection map and camera image. The cross-modal difference between the camera image and the projection map will make it difficult for the feature matching. Additionally, the scarcity of positive samples caused by the limited overlapping region will lead to the network's supervision problem.}
    \label{fig:2}
\end{figure}

Besides, the commonly used point cloud feature extraction network in the cross-modal registration task (e.g., PointNet++ \cite{qi2017pointnetpp}, sparse convolution network\cite{choy20194d}) always treats the LiDAR point cloud simply as a series of unordered points or voxels, which fails to effectively use the spatial and geometric characteristics of the point cloud. Therefore, to make better use of the scanning characteristics of LiDAR to bridge the domain gap, we introduce a method to project point clouds onto 2D representations: a range map and a reflectance map. Although using these projection methods is not new in other downstream tasks involving LiDAR point clouds, they have not been fully leveraged in cross-modal registration methods. Moreover, mapping point clouds to pixels in 2D representations offers a data format that implicitly contains more geometric information, making it more effective and accurate than using unordered sparse point clouds. This projection approach will enable more effective feature extraction and bridge the domain gap difference between point clouds and images. Furthermore, we combined the use of the two projection maps containing different types of information to enhance the system's robustness for registration across various scenarios.

After the projection, we transform the cross-modal registration task to an image registration task between the camera image and the projection maps. Common image registration methods are generally accomplished through key point matching. Due to the cross-modal differences between the projected 2D map and the camera image, traditional key point registration methods cannot be directly applied to this task, as the callenge 1 we showed in Fig. \ref{fig:2}. 

Furthermore, we previously proposed EEPNet \cite{yue2024eepnet}, which leverages point cloud projection to mitigate domain discrepancies between data of different dimensions. By matching edge pixels between the projected maps and camera images, it effectively reduces computational complexity. This method achieves highly accurate registration results on the KITTI dataset. Notably, due to the projection strategy and edge information utilization, the inference time is remarkably short, reaching as low as 15ms per frame. However, a limitation of this approach is that its training strategy appears effective only for dense point clouds, such as those in the KITTI dataset. When applied to sparse point clouds (e.g., nuScenes), the training loss fails to converge. Our analysis reveals that this issue stems from an insufficient proportion of matched edge pixels relative to the total pixels, which restricts EEPNet’s robustness and limits its applicability to broader scenarios, as shown in the callenge 2 of Fig. \ref{fig:2}.

To address these challenges, we design a multi-scale feature extraction network to mitigate the cross-modal differences between images and projection maps, effectively extracting features. Additionally, we introduce a patch-to-pixel feature matching framework. Specifically, patch-level searching addresses the supervision and success rate issues caused by positive sample scarcity, while pixel-level matching enhances registration accuracy. Experimental results show that the proposed method achieves high accuracy while maintaining low computational complexity. In summary, our contributions are as follows.
\begin{enumerate}
    \item  We propose an improved registration framework based on the previous work EEPNet \cite{yue2024eepnet} that leverages range map and reflectance map features for cross-modal registration between LiDAR point clouds and camera images, which can achieve a registration accuracy rate of over 99\% in the KITTI dataset. 
    \item  To address the cross-modal differences between the projection maps and the camera images in the current task, we propose a multi-scale feature extraction network that efficiently extracts features from both the images and the two different types of projection maps simultaneously.
    \item  We introduce a patch-to-pixel feature matching framework, which not only resolves the supervision problem that arise from the limited overlap between the projection map and the camera image but also enhances registration accuracy.
\end{enumerate}

\section{Related Work}

\subsection{Off-line Calibration Methods}
Traditional offline extrinsic calibration methods \cite{kim2019extrinsic,zhou2018automatic} are typically conducted before system operation, using manually placed prominent markers to identify 2D-3D correspondences and then estimate the external parameters between sensors using the least-square method \cite{zhang2004extrinsic}. These markers, such as checkerboard, generally contain distinct features on points \cite{geiger2012automatic}, lines \cite{bai2020lidar}, or planes \cite{zaiter2020extrinsic}. After establishing the correspondence between 2D and 3D feature points \cite{vali2018evaluation, lou2014image}. Such methods are performed before system deployment and can achieve high calibration accuracy. However, the marker placement and calibration process often require significant manpower and resources. If the sensor system's relative poses are actively changed or vibrations or shocks during operation cause sensor position shifts, the calibration needs to be repeated, which is time-consuming and labor-intensive.
\subsection{Online Calibration Methods}
Subsequently, many online calibration methods were proposed to optimize minor errors that occur during system operation, such as those caused by vibrations, after initial offline calibration. Existing approaches include using MLPs to estimate deviations \cite{yuan2020rggnet,iyer2018calibnet, lv2021lccnet}, keypoint-based methods\cite{ye2021keypoint}, and techniques that leverage objects \cite{sun2022atop} or segmentation \cite{zhu2020online, wang2020soic}. These real-time optimization techniques improve data alignment and enhance data fusion, leading to better performance in downstream tasks. However, these methods rely on the initial extrinsic parameters obtained from prior calibration and lose effectiveness when there are significant deviations between the actual and initial parameters. Moreover, deploying online calibration on top of complex offline calibration processes further increases costs.

\subsection{Cross-modal registration Methods}

Cross-modal registration methods are often used in visual localization tasks \cite{brachmann2017dsac, sarlin2019coarse, zhou2022geometry}, where the goal is to estimate the camera extrinsic parameters in the world coordinate system defined by large-scale point clouds. Traditionally, these methods are based on 2D visual descriptors \cite{detone2018superpoint, sarlin2020superglue, xie2024rcvs} to achieve alignment. Later advancements have introduced cross-dimensional descriptors \cite{feng20192d3d, pham2020lcd, wang2021p2} designed for direct cross-modal matching. These methods generally involve extracting features from different modalities to create high-dimensional representations, using feature similarity in this space to establish correspondences. Subsequently, transformation matrices are estimated using techniques like EPnP \cite{lepetit2009ep}.

For matching LiDAR point clouds and images, approaches like DeepI2P \cite{li2021deepi2p} reformulates the matching task into a classification problem to tackle cross-modal challenges. CorrI2P \cite{ren2022corri2p} later improves registration performance by leveraging overlap region extraction and feature exchange modules. VP2P-Match \cite{zhou2024differentiable} identifies the inherent differences between MLP networks \cite{qi2017pointnet++} used for point cloud feature extraction and CNNs used for image processing. By employing voxel sparse convolution, it reduced feature discrepancies and achieved better performance. However, these methods often treat point clouds as unordered sets of points or voxels, failing to fully use the geometric characteristics of LiDAR point clouds. 
\begin{figure}[htbp]
    \centering
    \includegraphics[width=0.97\linewidth]{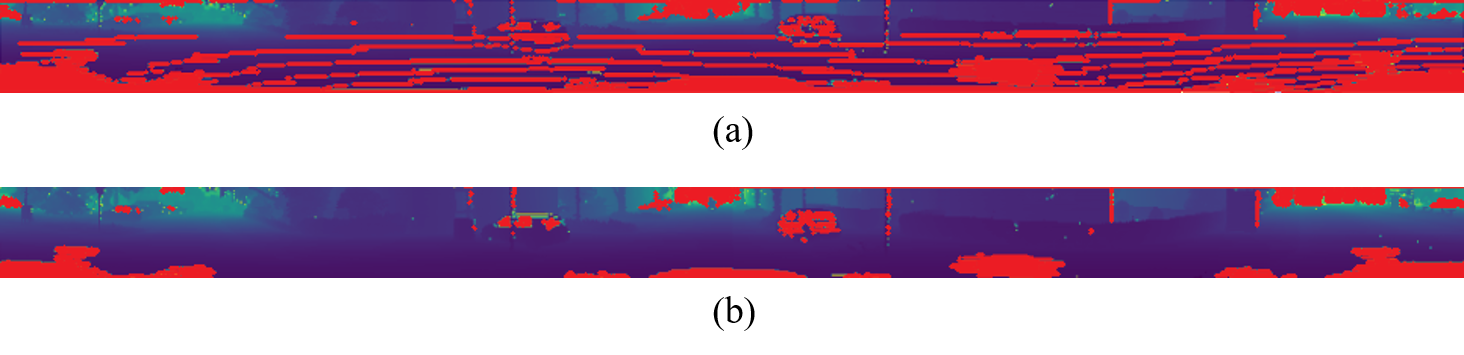}
    \caption{Visualizations of the projection map, (a) Projected map by spherical projection. The red area represents the empty holes. (b) Projected map from modified projection method using LaserID. The red area represents the empty holes. }
    \label{fig:3-2}
\end{figure}
\begin{figure}
    \centering
    \includegraphics[width=0.9\linewidth]{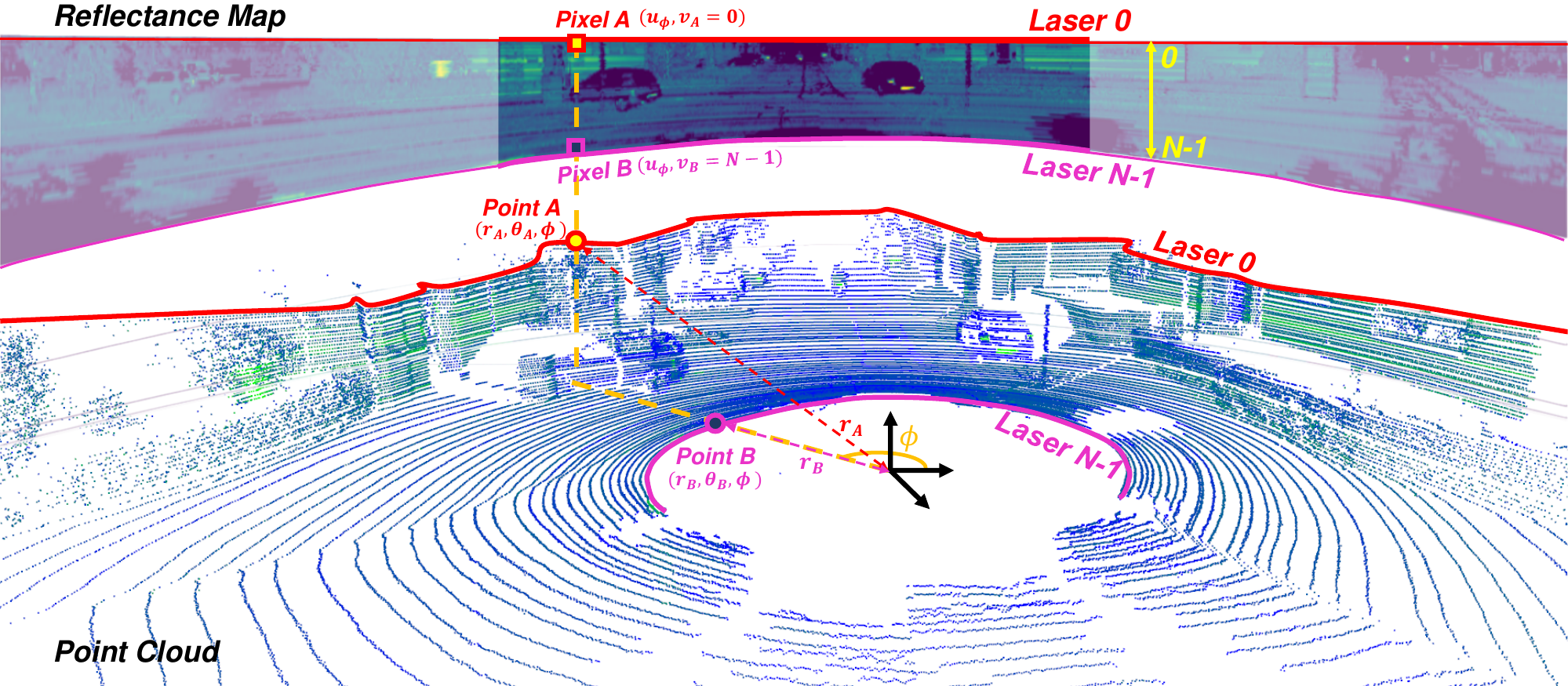}
    \caption{The detailed process of modified projection method using LaserID. The upper part of the image shows a projection map (exemplified by a reflectivity map), while the lower part displays the original point cloud being projected. Points A and B represent the 3D points captured by the uppermost and lowermost individual laser beams at the same azimuth angle, respectively. Pixels A and B correspond to the top and bottom pixels in the reflectivity map at the same horizontal coordinate.}
    \label{fig:3-1}
\end{figure}

\section{Proposed Method}
\subsection{Problem Statement}
Given an RGB image \( \mathbf{I}_C \in \mathbb{N}^{H \times W \times 3} \) and a point cloud \( \mathbf{P}_L \in \mathbb{R}^{N \times 4} \), where the four dimensions of the point cloud represent the spatial coordinates \( (x, y, z) \) and the reflectance intensity \(r\), respectively, our goal is to estimate the rigid transformation matrix \(\mathbf{T}\) to align the 3D points to the 2D image coordinates. Matrix \(\mathbf{T}\) is given by a translation vector \(\mathbf{t} \in \mathbb{R}^3\) and a rotation matrix \(\mathbf{R} \in \text{SO}(3)\) as follow:

\[
\mathbf{T} = \begin{bmatrix}
\mathbf{R} & \mathbf{t} \\
\mathbf{0} & 1
\end{bmatrix} \in \text{SE}(3).
\tag{1}
\]

\subsection{Data Pre-processing}

To get a better use of the geometric characteristics of LiDAR point cloud, and mitigate the impact of domain gap between the point cloud and the image, we project the original 3D point cloud into 2D representations by a specific projection method. Currently, there are various methods for such projection, such as spherical projection \cite{chen2017multi}, cylindrical projection \cite{milioto2019rangenet++}, and bird's-eye view (BEV) projection \cite{wu2021detailed}. Among them, the spherical projection can make good use of the characteristics of lateral scanning of LiDAR. However, according to the research by Wu \textit{et al.} \cite{wu2021detailed}, direct spherical projection may result in numerous empty holes in the projected map, due to slight misalignments between the coordinate systems of individual lasers and the overall LiDAR system, as shown in the red area of Figure \ref{fig:3-2}(a). These empty areas could negatively impact subsequent registration performance \cite{xu2024igreg}. 

To address this issue, we use a modified the spherical projection by associating the vertical axis of the projection map with the LaserID of the LiDAR point cloud instead of the elevation angle, which can reduce the empty holes effectively, as illustrated in Figure \ref{fig:3-2}(b). By using the radius in the spherical coordinate system as the information projected onto the map, we obtain the range map \( \mathbf{I}_{Ra}\in \mathbb{N}^{H_R \times W_R} \). Similarly, projecting the reflectance intensity yields the reflectance map \( \mathbf{I}_{Rf}\in \mathbb{N}^{H_R \times W_R}  \). These maps, along with the camera image \( \mathbf{I}_C \in \mathbb{N}^{H \times W \times 3} \), are fed into the subsequent network. 

To explain this projection process vividly, we take the reflectance map \(\mathbf{I}_{Rf}\) as an example, as shown in Figure \ref{fig:3-1}. For an \( N \)-line LiDAR, because the horizontal axis of the projection map corresponds to the azimuth angle \( \phi \) in the spherical coordinate system and the vertical axis corresponds to the LaserID, we can assume that we have two data points at the same azimuth angle \( \phi \): point A in the first line (LaserID = 0) with spherical coordinates \( (r_A, \theta_A, \phi) \) and point B in the last line (LaserID = \( N-1 \)) with coordinates \( (r_B, \theta_B, \phi) \), where, \( r_A \) and \( r_B \) are the spherical radii of points A and B, and \( \theta_A \) and \( \theta_B \) are their pitch angles. After this LaserID-based projection, they are mapped to the same column \( u_\phi \), with Pixel A at \( (u_\phi, 0) \) in the first row and Pixel B at \( (u_\phi, N-1) \) in the last row of the reflectance map \cite{wu2023accelerating}.
\begin{figure*}[htbp]
    \centering
    \includegraphics[width=0.85\linewidth]{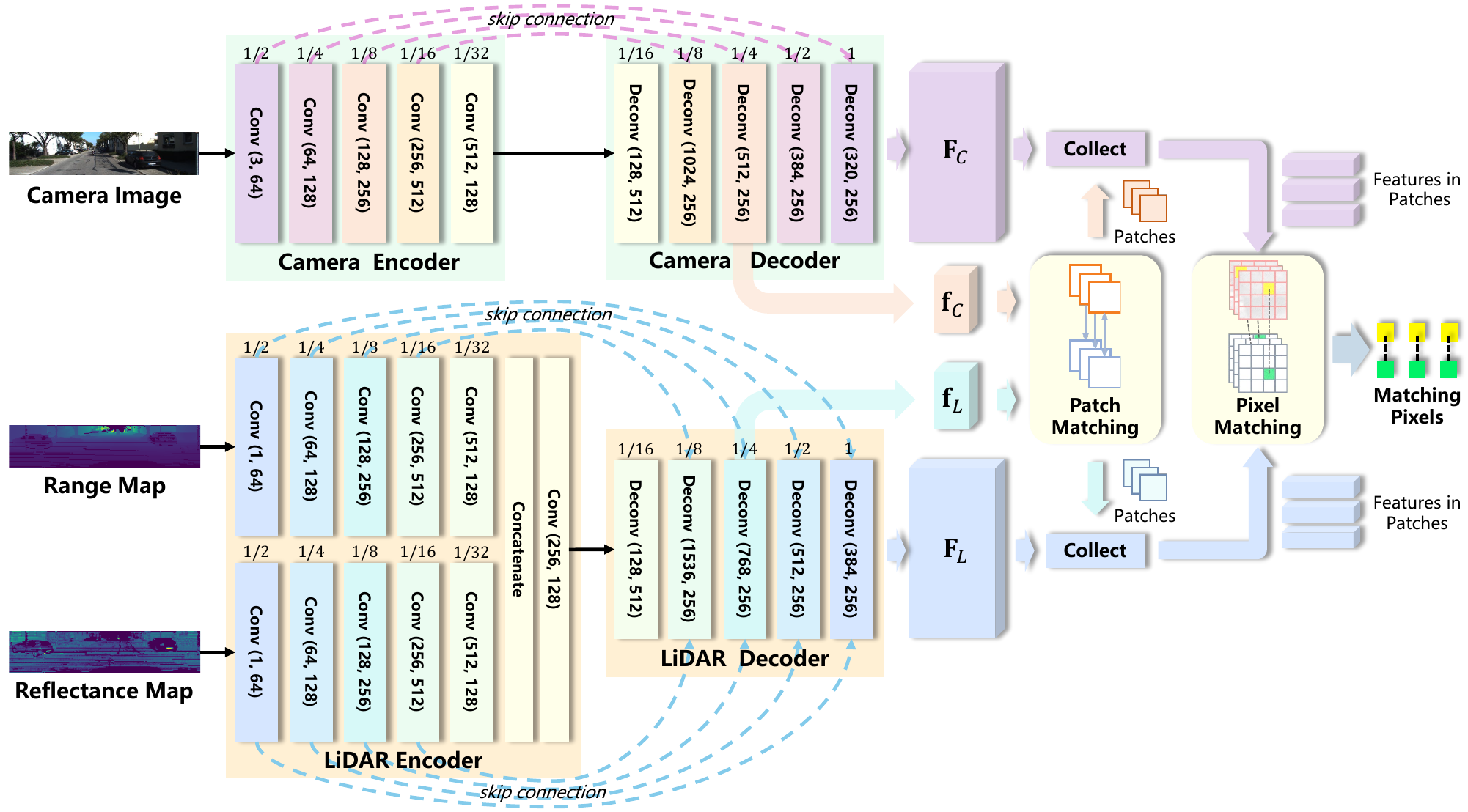}
    \caption{The overall network architecture of the proposed method, illustrates the details of the feature extraction network and the overview of the patch-to-pixel matching process. Specifically, the feature extraction network mainly consists of four parts: Camera Encoder, Camera Decoder, LiDAR Encoder, and LiDAR Decoder.}
    \label{fig:4}
\end{figure*}

\subsection{Feature Extraction Network}
Unlike other image registration tasks, the cross-modal difference between the projection map and the camera image require a specialized image feature extraction network. Additionally, to address the point cloud degradation problem within the camera's field of view in some scenarios, we have added an extra branch for extracting reflectance features from \(\mathbf{I}_{Rf}\) alongside the spatial features from \( \mathbf{I}_{Ra} \). This enhancement effectively increases the system's robustness across different environments. The details of the feature extraction network architecture can be seen on the left side of Figure \ref{fig:4}.
\begin{figure*}[htbp]
    \centering
    \includegraphics[width=0.85\linewidth]{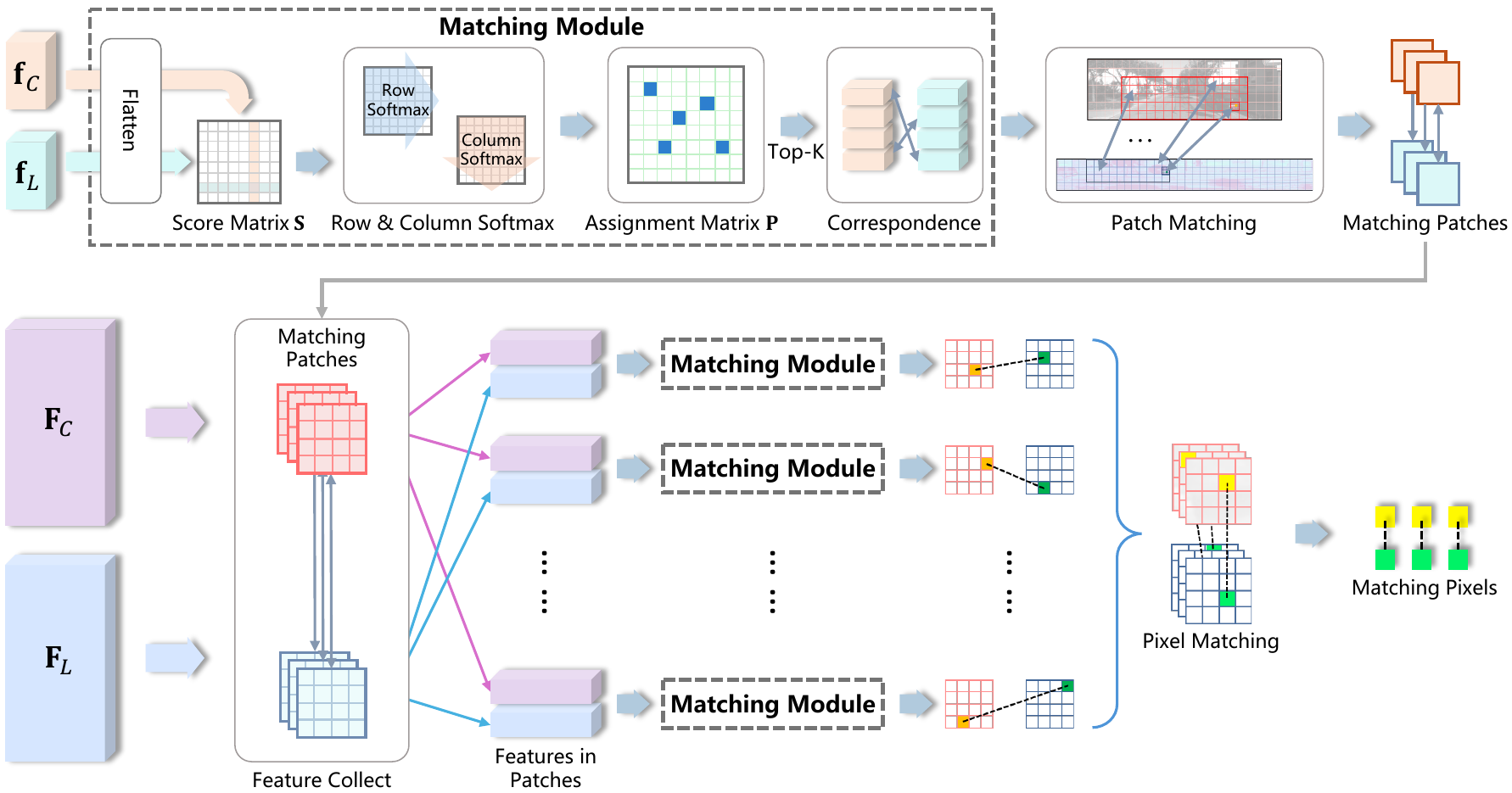}
    \caption{Detailed architecture of the patch-to-pixel matching network. The upper half depicts the patch matching network, while the lower half illustrates the pixel matching part. The gray dashed box highlights the Matching Module, which shares an identical structure in both part.}
    \label{fig:5}
\end{figure*}
We employ convolutional neural networks (CNNs) to process the camera image, range map, and reflectance map. The camera encoder extracts features from \( \mathbf{I}_C \), producing a high-dimensional feature at 1/32 of the original image scale. This small-scale, high-dimensional feature is then passed to the camera decoder. In the camera decoder network, the feature scale is gradually upsampled. First, after three deconvolution layers, we obtain the patch-level feature at 1/4 of the original image scale, \( \mathbf{f}_C \in \mathbb{R}^{\frac{H}{4} \times \frac{W}{4} \times D_{\text{patch}}} \), where \(D_{patch}\) is the number of patch-level feature channels. Then after two layers of deconvolution to achieve pixel-level features \( \mathbf{F}_C \in \mathbb{R}^{H \times W \times D_{\text{pixel}}} \) at the original image scale, where \(D_{pixel}\) is the number of pixel-level feature channels.

For the LiDAR encoder, which processes the projection maps, we use the same CNN architecture but with two identical encoder branches for the range map and reflectance map, respectively. The features extracted from these branches are concatenated and passed through the LiDAR decoder. The LiDAR decoder then outputs patch-level features \( \mathbf{f}_L \in \mathbb{R}^{\frac{H_R}{4} \times \frac{W_R}{4} \times D_{\text{patch}}} \), and subsequently, two layers of transposed convolution are applied to obtain pixel-level features \( \mathbf{F}_L \in \mathbb{R}^{H_R \times W_R \times D_{\text{pixel}}} \) at the original image scale.
    
\subsection{Patch-to-Pixel Matching Network}

To address the issue of positive sample scarcity in feature point matching caused by the limitation of the overlapping region between the camera image and the point cloud. We first conduct searches at the patch level, followed by pixel-level matching within the matched patches. This matching strategy not only solves the network supervision problem by increasing the proportion of positive samples through the patch-level searching but also significantly enhances registration accuracy through the pixel-level matching.

The structure of our matching network is shown in Figure \ref{fig:5}. Initially, we input the patch-level features \( \mathbf{f}_C \) and \( \mathbf{f}_L \), extracted by the feature extraction network, into a matching module to perform patch-level matching. The matching module is similar to the optimization network structure in Lightglue \cite{lindenberger2023lightglue}. First, we compute the score matrix \( \mathbf{S} \) between \( \mathbf{f}_C \) and \( \mathbf{f}_L \). Specifically, these features are flattened and then passed through a linear layer before performing matrix multiplication
\[
\mathbf{S} = \text{Linear}\left(\text{Flatten}\left(\mathbf{f}_{C}\right)\right) \times \text{Linear}\left(\text{Flatten}\left(\mathbf{f}_{L}\right)\right)^\mathsf{T}.
\tag{2}
\]

Next, we apply softmax operations on \( \mathbf{S} \) along both rows and columns, after computing the Hadamard product of the results we can get the soft assignment matrix \(\mathbf{P}\),
\[
\mathbf{P}_{ij} = \frac{\exp(\mathbf{S}_{ij})}{\sum^{N_{2D}}_{k =1} \exp(\mathbf{S}_{kj})} \cdot \frac{\exp(\mathbf{S}_{ij})}{\sum^{N_{3D}}_{k=1} \exp(\mathbf{S}_{ik})},\tag{3}
\]where \(\mathbf{P}_{ij} \) represents the value at the \( i \)-th row and \( j \)-th column of the assignment matrix \( \mathbf{P} \), and \( \mathbf{S}_{ij} \) represents the value at the \( i \)-th row and \( j \)-th column of the assignment matrix \( \mathbf{S} \). We select the top-k entries from the resulting matrix to establish matches between patches based on feature similarity. Since the patch feature scale is 1/4 of the original image size, each patch corresponds to a 4×4 image region, which includes 16 pixels. 

We then collect the pixel-level features from the matched patches using \(\mathbf {F}_C \) and \( \mathbf{F}_L \), which are at the original image scale, and then we use them for pixel-level matching. This matching is done for each individual patch pair, with all pixels within each patch pair passed into another matching module, identical to the one used for patch matching, except that this module only retains the single highest similarity value (top-1). In other words, within each matching patch pair, we use the matching module to select one pair of matching pixels. By applying this process to each matching patch, we obtain the final matched pixels.

With these matched pixels, we can establish the 3D-2D correspondences between the projected 3D point cloud points and the camera image pixels. Subsequently, we use the EPnP \cite{lepetit2009ep} method with RANSAC \cite{fischler1981random} to solve for the pose, ultimately estimating the transformation matrix between the LiDAR point cloud and the camera image. 

\subsection{Loss Function}
In the loss function design, we first project the LiDAR point cloud onto the camera image using the ground truth extrinsic parameters from the dataset to obtain the actual 3D-2D correspondences. Assuming we have the actual correspondences \( \{\mathcal{C}\} \), containing \( M \) correspondences, we divide each value in \(\mathcal{C} \) by 4 and take the floor to derive the ground truth patch correspondences \( \{\mathcal{C}^* \}\). For each ground truth patch correspondence, its corresponding value in the patch matching matrix \( \mathbf{P} \) should be as large as possible. Based on this idea, we formulate the first part of our loss function, \(\mathcal{ L}_{\text{patch}} \), 

\[\mathcal{L}_{patch} = -\frac{1}{|{\mathcal{C}^*}|} \sum_{(i,j)\in {\mathcal{C}^*}} \log{\mathbf{P}_{ij}}.\tag{4}\]

During training, we do not use top-k in the matching network, instead, we use the ground truth patch correspondences for subsequent pixel-level matching directly. For each patch correspondence, the network calculates \(M\) assignment matrices \( \mathbf{p} \). By dividing the actual correspondences \(\mathcal{C}\)  by 4 and taking the remainder, we obtain the ground truth pixel correspondences \(\{ \hat{\mathcal{C} }\} \) in each \(\mathbf{p} \) matrix, which leads to the second part of our loss function, \( \mathcal{L}_{\text{pixel}} \):
\[\mathcal{L}_{pixel} = -\frac{1}{{M}} \sum_{i=1}^{M} \log{\mathbf{p}^i_{\hat{\mathcal{C}_i}}},\tag{5}\]where \( \mathbf{p}^i \) denotes the assignment matrix computed for the \( i \)-th pair of patches in the matching module, \( \hat{\mathcal{C}_i} \) represents the \( i \)-th ground truth pixel correspondence, and \( \mathbf{p}^i_{\hat{\mathcal{C}_i}} \) refers to the value at the corresponding position in \( \mathbf{p}^i \).

During training, we supervise both parts jointly, resulting in a total loss function that is the sum of the two components, expressed as

\[
\mathcal{L}_{\text{total}} = \mathcal{L}_{\text{patch}} + \mathcal{L}_{\text{pixel}}.\tag{6}\
\]

\section{Experimental Results and Analysis}
\begin{table*}[htbp]
\centering
\caption{Registration accuracy on the KITTI and nuScenes datasets. Lower is better for RTE and RRE, higher is better for Acc.}
\label{tab:t1}
\begin{tabular}{>{\raggedright\arraybackslash}p{0.15\linewidth}|>{\centering\arraybackslash}p{0.1\linewidth}>{\centering\arraybackslash}p{0.1\linewidth}>{\centering\arraybackslash}p{0.07\linewidth}|>{\centering\arraybackslash}p{0.1\linewidth}>{\centering\arraybackslash}p{0.1\linewidth}>{\centering\arraybackslash}p{0.07\linewidth}}
\hline
\multirow{2}{*}{Method} &
  \multicolumn{3}{c|}{KITTI} &
  \multicolumn{3}{c}{nuScenes} \\ \cline{2-7} 
 &
  \multicolumn{1}{c|}{RTE(m)↓} &
  \multicolumn{1}{c|}{RRE(°)↓} &
  Acc.↑ &
  \multicolumn{1}{c|}{RTE(m)↓} &
  \multicolumn{1}{c|}{RRE(°)↓} &
  Acc.↑ \\ \hline
Grid Cls. + PnP 
 \cite{li2021deepi2p}&
  \multicolumn{1}{c|}{3.64 ± 3.46} &
  \multicolumn{1}{c|}{19.19 ± 28.96} &
  11.22 &
  \multicolumn{1}{c|}{3.02 ± 2.40} &
  \multicolumn{1}{c|}{12.66 ± 21.01} &
  2.45 \\
DeepI2P (3D) 
 \cite{li2021deepi2p}&
  \multicolumn{1}{c|}{4.06 ± 3.54} &
  \multicolumn{1}{c|}{24.73 ± 31.69} &
  3.77 &
  \multicolumn{1}{c|}{2.88 ± 2.12} &
  \multicolumn{1}{c|}{20.65 ± 12.24} &
  2.26 \\
DeepI2P(2D) 
 \cite{li2021deepi2p}&
  \multicolumn{1}{c|}{3.59 ± 3.21} &
  \multicolumn{1}{c|}{11.66 ± 18.16} &
  25.95 &
  \multicolumn{1}{c|}{2.78 ± 1.99} &
  \multicolumn{1}{c|}{4.80 ± 6.21} &
  38.10 \\
CorrI2P 
 \cite{ren2022corri2p}&
  \multicolumn{1}{c|}{3.78 ± 65.16} &
  \multicolumn{1}{c|}{5.89 ± 20.34} &
  72.42 &
  \multicolumn{1}{c|}{3.04 ± 60.76} &
  \multicolumn{1}{c|}{3.73 ± 9.03} &
  49.00 \\ 
VP2P-Match \cite{zhou2024differentiable}&
  \multicolumn{1}{c|}{0.75 ± 1.13} &
  \multicolumn{1}{c|}{3.29 ± 7.99} &
  83.04 &
  \multicolumn{1}{c|}{0.89 ± 1.44} &
  \multicolumn{1}{c|}{2.15 ± 7.03} &
  88.33 \\ 
  EEPNet\cite{yue2024eepnet}&
  \multicolumn{1}{c|}{0.54 ± 0.85} &
  \multicolumn{1}{c|}{2.97 ± 3.05} &
  85.74&
  \multicolumn{1}{c|}{-} &
  \multicolumn{1}{c|}{-} &
  -\\ 
EEPNet-V2 &
  \multicolumn{1}{c|}{\textbf{0.21 ± 0.25}} &
  \multicolumn{1}{c|}{\textbf{0.67 ± 0.80}} &
  \textbf{99.03}&
  \multicolumn{1}{c|}{\textbf{0.82 ± 0.74}} &
  \multicolumn{1}{c|}{\textbf{0.87 ± 2.12}} &
  \textbf{91.50} \\ \hline
\end{tabular}
\begin{tablenotes}
\footnotesize
\item[*]           *  The vacant data in the table indicates that the method could not be effectively trained on this dataset.
\end{tablenotes}
\end{table*}
\subsection{Dataset}
We conducted experiments on the KITTI Odometry and nuScenes datasets. 

\textbf{KITTI Odometry\cite{geiger2012we}.} We generated image-point cloud pairs from the same data frame of 2D/3D sensors. Following previous works, we used sequences 00-08 for training and sequences 09-10 for testing. We downsample the camera image resolution to 512×160 and employ reflectance maps with a resolution of 1024×64 for both training and testing. Artificial errors applied to the point clouds during training and testing include ±10m two-dimensional translations along the \textit{x} and \textit{y} axes and arbitrary rotations around the \textit{z}-axis. It is worth noting that, since the center position in the spherical coordinates of point clouds is typically fixed, translations along the \textit{x} and \textit{y} axes applied in other works do not accurately reflect real-world scenarios. To ensure fairness and realism, we initially applied arbitrary rotations around the \textit{z}-axis to the point cloud before generating reflectance maps. then we applied translations along the \textit{x} and \textit{y} axes to the point cloud. 

\textbf{nuScenes\cite{caesar2020nuscenes}.} For nuScenes, we used 850 scenes for training and 150 scenes for testing. Additionally, we downsample the camera image resolution to 320×160 and employ reflectance maps with a resolution of 1024×32 for both training and testing. Furthermore, we applied random errors ±10m two-dimensional translations along the \textit{x} and \textit{y} axes and arbitrary rotations around the \textit{z}-axis to the nuScenes dataset, using the same method as KITTI.
\begin{figure*}[htbp]
    \centering
    \includegraphics[width=0.85\linewidth]{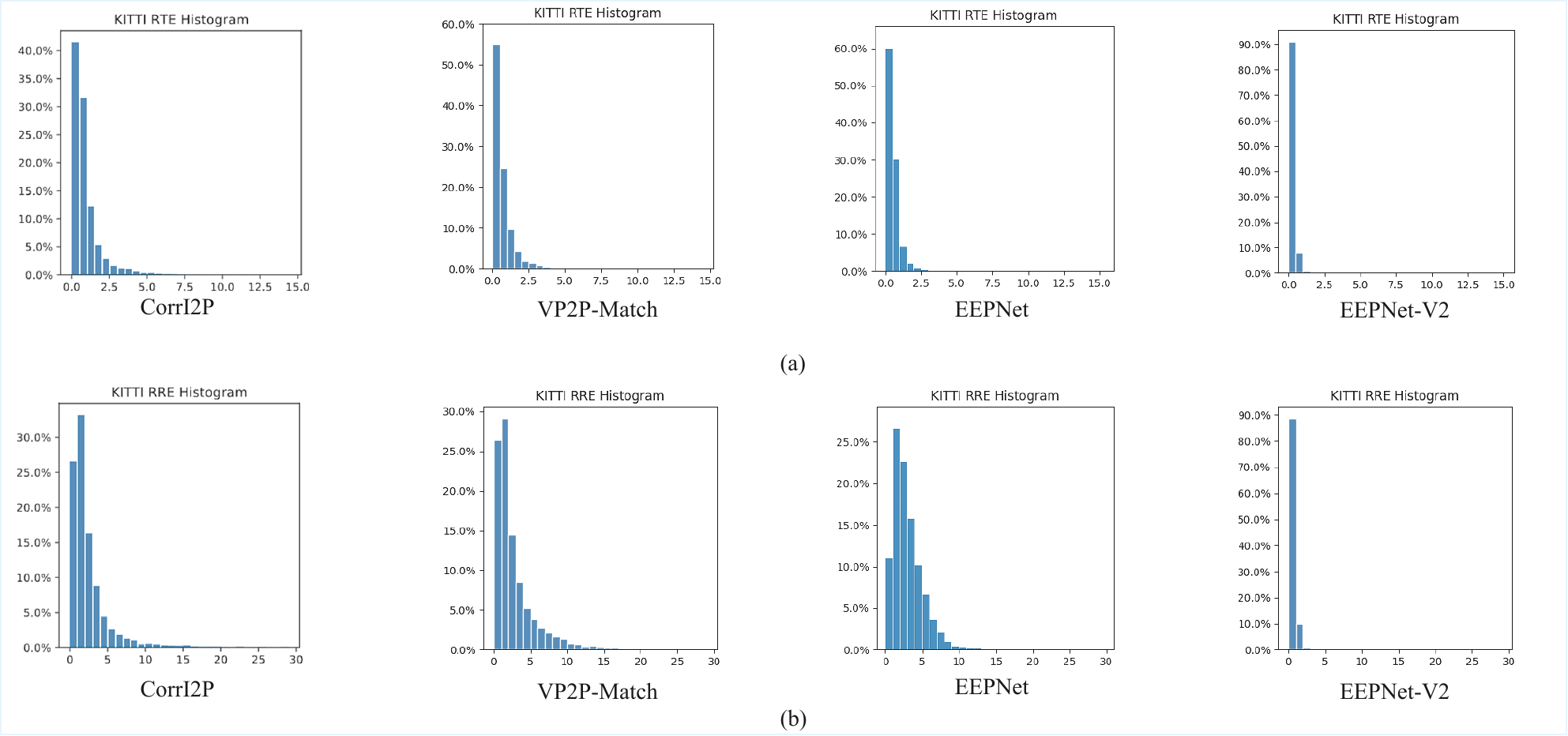}
    \caption{Histograms of distributions of RTE and RRE on the KITTI dataset. x-axis is RTE(m) and RRE(°), and y-axis is the percentage. (a) Histogram for RTE (b) Histogram for RRE}
    \label{fig:6}
\end{figure*}
\subsection{Compared Methods}
We benchmarked the proposed method against three methods: DeepI2P \cite{li2021deepi2p}, CorrI2P \cite{ren2022corri2p}, and VP2P-Match \cite{zhou2024differentiable}. 
\begin{enumerate}
    \item \textbf{DeepI2P.} This approach offers two distinct methods: \textbf{Grid Cls. + PnP} and \textbf{Frus. Cls. + Inv. Proj.}. The \textbf{Grid Cls. + PnP} method first segments the input image into a 32×32 grid. Then, it trains a neural network to classify 3D points into specific 2D grid cells. Finally, it uses the EPnP algorithm and RANSAC to estimate the transformation matrix \(\mathbf{T}\). The \textbf{Frus. Cls. + Inv. Proj.} method introduces frustum classification using inverse camera projection to determine the transformation matrix, exploring both 2D and 3D inverse projections, referred to as DeepI2P(2D) and DeepI2P(3D), respectively.
    \item \textbf{CorrI2P.} Building upon DeepI2P, CorrI2P enhances registration accuracy by using a neural network equipped with a feature exchange module and supervision of overlapping regions. This method learns correspondences between image and point cloud pairs.
    \item \textbf{VP2P-Match.} VP2P-Match focuses on improving the accuracy and speed of pixel-to-point matching by leveraging sparse convolution to enhance the similarity between point cloud features and CNN-extracted image features. It incorporates a differentiable PnP solver into an end-to-end training framework, which provides a better registration performance.
\end{enumerate}

\subsection{Registration Accuracy}
To evaluate registration accuracy and success rate, we used the same statistical methods as in VP2P-Match\cite{zhou2024differentiable}. Specifically, we used the relative translational error (RTE) \(E_t\) and the relative rotational error (RRE) \(E_R\) to evaluate our registration results. These errors are computed as

\[E_R = \sum_{i=1}^{3} \left| \gamma(i) \right|,\tag{7}\]
\[E_t = \left\| \mathbf{t}_{gt} - \mathbf{t}_E \right\|,\tag{8}\]where \(\gamma(i), i=1,2,3\) are the Euler angles of the matrix \(\mathbf{R}_{gt}^{-1}\mathbf{R}_E\). Here, the rotation matrix \(\mathbf{R}_{gt}\) and the translation vector \(\mathbf{t}_{gt}\) denote the transformation of the ground truth, while matrix \(\mathbf{R}_E\) and vector \(\mathbf{t}_E\) define the estimated transformation. We calculated the mean and standard deviation of all the data without removing the image-point cloud samples with large errors, as done in CorrI2P. Furthermore, for the registration success rate (Acc.), registrations with \(\text{RTE} < 2\space m\) and \(\text{RRE} < 5\)° are considered successful. The final results are presented in Table \ref{tab:t1}.

At the same time, we compared the distribution of RTE and RRE in different intervals with other methods using histograms in Figure \ref{fig:6}. We conducted the comparison on the KITTI dataset. A higher proportion in the leftward intervals indicates smaller errors. As can be seen, our method achieves around 90\% of matching cases where \(\text{RTE} < 0.5\space m\) and \(\text{RRE} < 1\)°, significantly surpassing the other methods, which demonstrates the accuracy of the proposed method.

\subsection{Efficiency Comparison}
\begin{table}[htbp]
\centering
\caption{The efficiency comparisons with other methods.}
\label{tab:t2}
\begin{tabular}{c|c|c}
\hline
\multicolumn{1}{l|}{} & Model Size(MB)& Inference Time(s)\\ \hline
DeepI2P(2D)           & 100.12         & 23.47             \\
DeepI2P(3D)           & 100.12         & 35.61             \\
CorrI2P               & 141.07         & 8.96              \\
VP2P-Match            & \textbf{30.73} & 0.19             \\
EEPNet                & 58.84          & \textbf{0.015}\\ 
EEPNet-V2             & 36.09          & {0.12}\\ \hline
\end{tabular}
\end{table}
We compared the model size and inference time of our method with other methods on a platform equipped with an NVIDIA GeForce RTX 3090 GPU and an AMD Ryzen Threadripper 2990WX. The results are shown in Table \ref{tab:t2}. We evaluated our conducted efficiency test on the KITTI dataset, calculating the inference time per frame by dividing the total inference time with the total number of input frames. From the results of the efficiency comparison, we can see that our model has a relatively low number of parameters and short runtime, allowing the registration process to be real-time. 

\subsection{Visualization Result}
\begin{figure*}[htbp]
    \centering
    \includegraphics[width=0.95\linewidth]{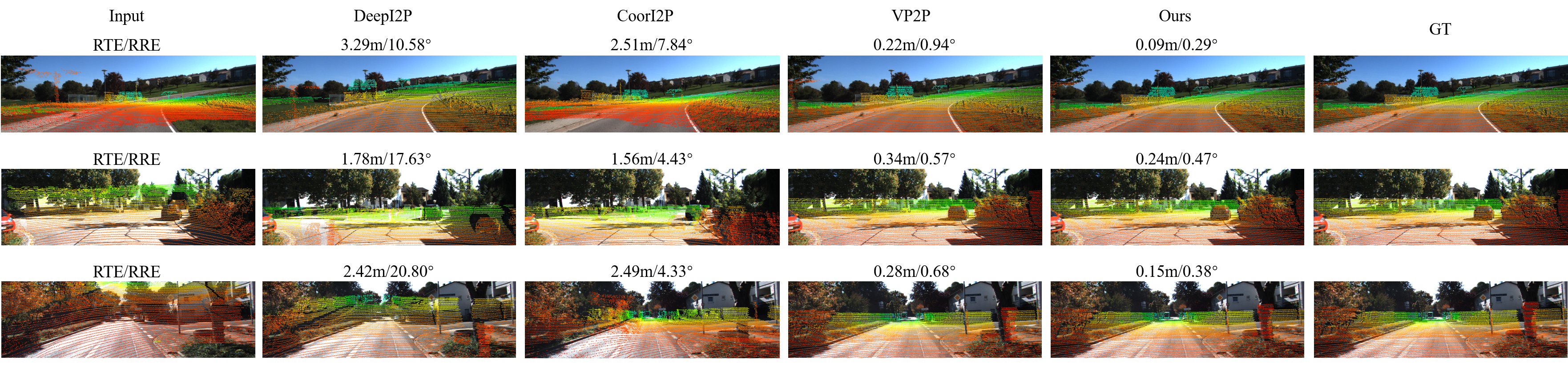}
    \caption{Visualization of registration results, demonstrating the projection of LiDAR point clouds onto the camera plane after transformation using different extrinsic calibration methods. The color of projected points indicates their distance from the sensor (range), with warmer hues representing closer points and cooler hues representing farther points.}
    \label{fig:8}
\end{figure*}

Fig. \ref{fig:8} presents a comparison of the registration results between different methods on the KITTI dataset, displaying the projections of the point clouds in the image plane. The color of the points in the images represents the distance from the camera. We show the initial unaligned projection results (Input), the aligned projection results obtained using the ground truth (GT), and the projection results after registration using different registration methods. It shows that our method achieves the the most accurate registration result in different environments.

\begin{figure}[htbp]
    \centering
    \includegraphics[width=0.8\linewidth]{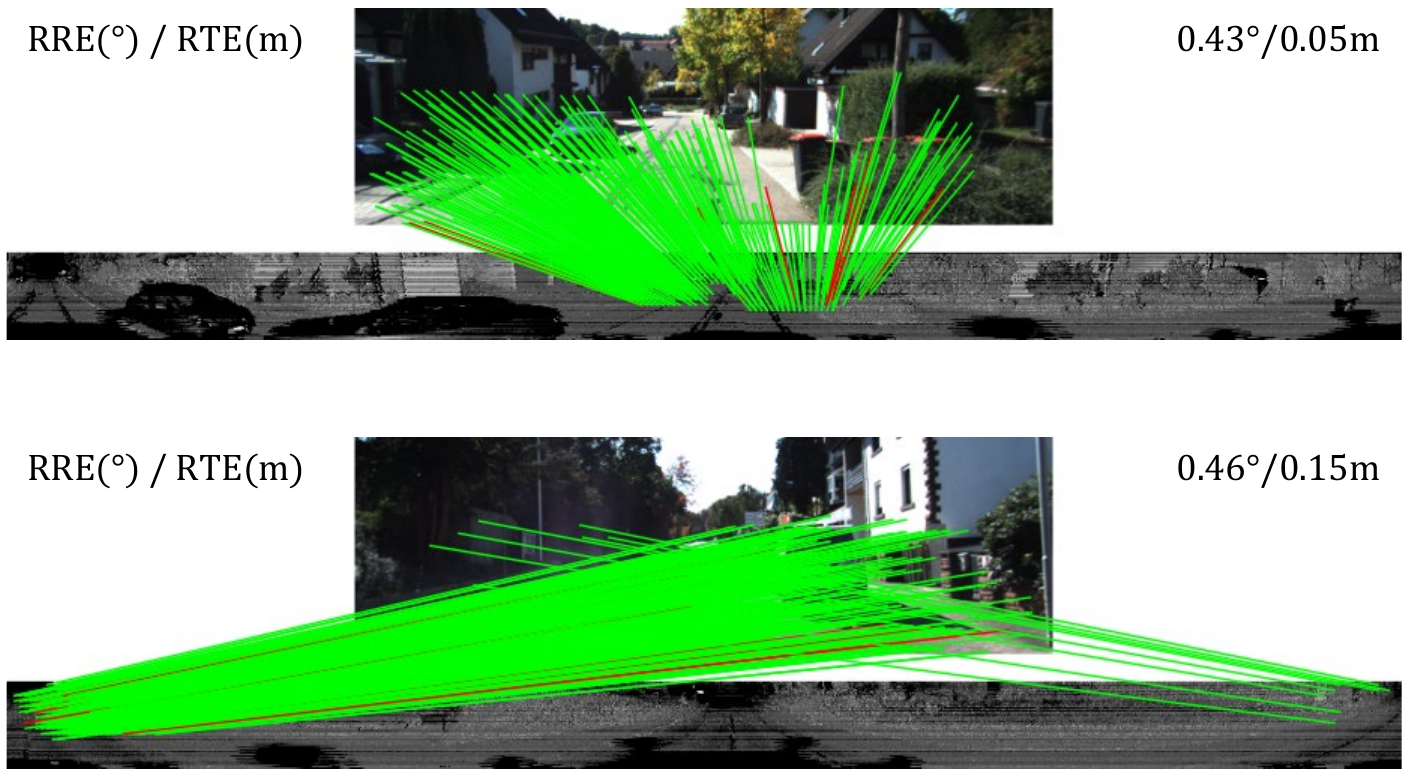}
    \caption{Visualization of the matching results. The image above is the camera image, and the one below is the reflectance map. Outlier pairs discarded by RANSAC are indicated with red lines, and accurate matching pairs are connected with green lines.}
    \label{fig:7}
\end{figure}

Besides, we visualized the matching results in Fig. \ref{fig:7}. Specifically, we present the matching results between the camera image and the reflectance map. The estimated matching pixel pairs are connected by lines in this graph, which will be used for subsequent pose estimation. In the pose estimation process, outlier pairs discarded by RANSAC are marked with red lines, indicating inaccurate estimates, and the accurate matching pairs are connected by green lines. These images show that the estimated matching pairs are not only sufficiently numerous but also highly accurate. Notably, the second figure shows that even when the overlapping region of the point cloud and the image are projected onto the boundary area, the proposed method can still estimate the correspondences well, which prove the robustness of our method. 

\subsection{Ablation Study}
In the ablation study section, we validated the effectiveness of certain network components and examined the impact of different selections of top-k value on the network's performance. All the experiments are conducted on the KITTI dataset.

First, we primarily conduct ablation experiments on the range map and reflectance map branches in the feature extraction network to prove the effectiveness of the additional information. The results are shown in Table \ref{tab:t3}. This table shows that neither branch achieves the best accuracy without the additional information. In some scenarios where point cloud degradation occurs within the camera's field of view, relying solely on the range map’s point cloud geometric information fails to provide good registration. The additional reflectance information helps to address the issues in such scenes. Similarly, if only the reflectance map is used without the geometric information for assistance, it cannot handle some special cases where the reflectance information is not enough.

Second, we tested the impact of different selections of top-k value during the patch matching stage on registration accuracy and inference time, As shown in Table \ref{tab:t4}. When the top-k value is small, the accuracy slightly decreases, but the computation speed significantly improves. However, increasing the top-k value does not always lead to better performance. In fact, when the top-k value is too large, there is no significant improvement in accuracy, it results in a substantial decrease in efficiency. Therefore, in practical use, it is important to choose an appropriate value that balances both accuracy and speed.

\begin{table}[htbp]
\centering
\caption{Experimental results on ablation of feature extraction network branches}
\label{tab:t3}

\begin{tabular}{l|c|c|c}
\hline
                 & RTE(m)↓              & RRE(°)↓              & Acc.↑          \\ \hline
reflectance only & 0.29 ± 0.59          & 0.99 ± 3.62          & 98.08          \\
range only       & 0.30 ± 1.23          & 1.08 ± 7.89          & 98.17          \\ \hline
Full             & \textbf{0.21 ± 0.25}& \textbf{0.67 ± 0.80}& \textbf{99.03}\\ \hline
\end{tabular}
\end{table}

\begin{table}[htbp]
\centering
\caption{Results of ablation experiments on top-k selection. \textquoteleft Time\textquoteright \space for the inference time.}
\label{tab:t4}
\begin{tabular}{c|c|c|c|c}
\hline
Top-k & RTE(m)↓     & RRE(°)↓     & Acc.↑ & Time(s)↓ \\ \hline
100   & 0.29 ± 0.86 & 0.86 ± 4.27 & 98.41 & 0.048    \\
200   & 0.24 ± 0.74 & 0.76 ± 4.03 & 99.01 & 0.086    \\
300   & 0.21 ± 0.25 & 0.67 ± 0.80 & 99.03 & 0.123    \\
400   & 0.21 ± 0.44 & 0.69 ± 2.79 & 99.18 & 0.161    \\
500   & 0.20 ± 0.56 & 0.69 ± 3.21 & 99.40 & 0.197    \\
600   & 0.19 ± 0.24 & 0.64 ± 0.56 & 99.34 & 0.233    \\ \hline
\end{tabular}
\end{table}

\section{Conclusion}
We propose a framework for cross-modal registration task by projecting 3D LiDAR point clouds into 2D representations. The projection method better leverages the geometric properties of the point clouds while reducing the impact of cross-modal differences. Furthermore, to address the limitation in the previous method EEPNet in the image matching process of the registration task, we introduce a multi-scale feature extraction network and a patch-to-pixel matching strategy to get better adaption. Experiments on the KITTI and nuScenes datasets demonstrate that our network achieves exceptional accuracy, precision, and robustness.

\small
\bibliographystyle{IEEEbib}
\bibliography{main}
\vspace{-3cm}
\begin{IEEEbiography}
[{\includegraphics[width=1in,height=1.25in,clip,keepaspectratio]{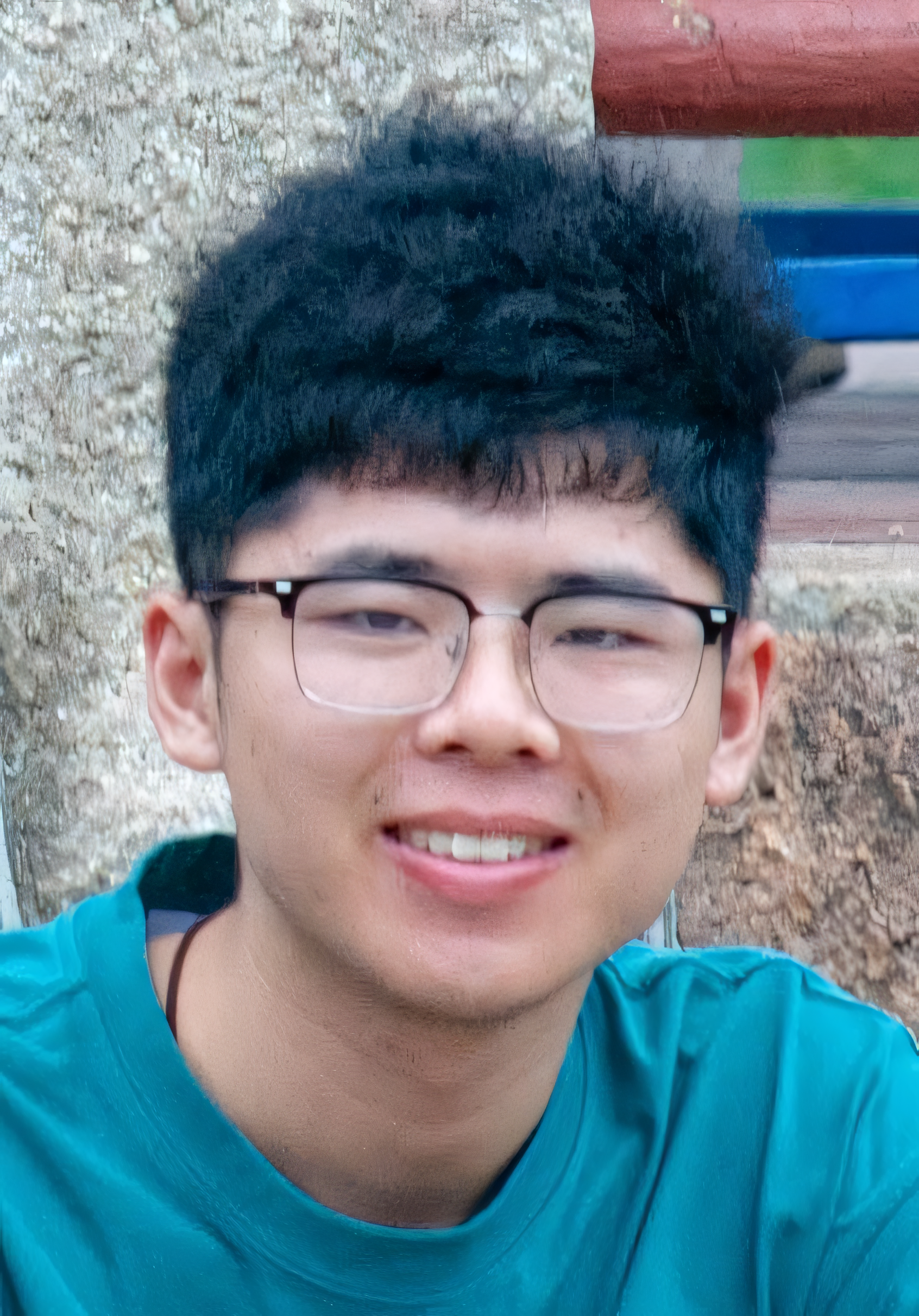}}] 
{Yuanchao Yue} received the B.S. degree in automation from Yanshan University, Hebei, China, in 2022. He is currently working toward the M.S. degree with the School of Control Science and Engineering, Shandong University, Jinan, China. His current research interests include deep learning, computer vision, cross-modal registration, point cloud registration and multi-sensor fusion.					                 
\end{IEEEbiography}
\vspace{-1cm}
\begin{IEEEbiography}
[{\includegraphics[width=1in,height=1.25in,clip,keepaspectratio]{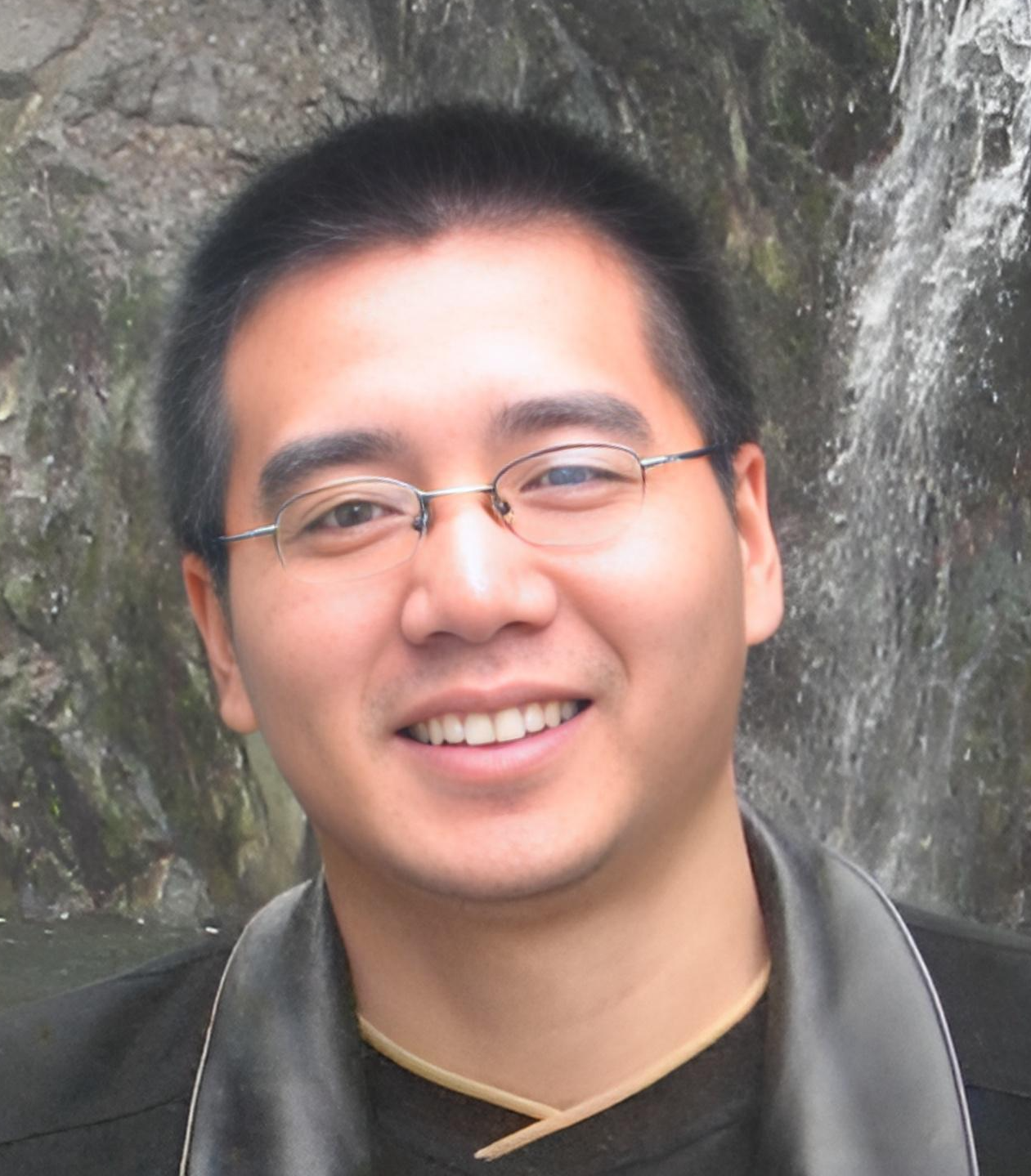}}] 
{Hui Yuan} (SeniorMember,IEEE) received the B.E. and Ph.D. degrees in telecommunication engineering from Xidian University, Xi’an, China, in 2006 and 2011, respectively. In April 2011, he joined Shandong University, Ji’nan, China, as a Lecturer (April 2011–December 2014), an Associate Professor (January 2015-October 2016), and a Professor (September 2016). From January 2013-December 2014, and November 2017-February 2018, he also worked as a Postdoctoral Fellow (Granted by the Hong Kong Scholar Project) and a Research Fellow, respectively, with the Department of Computer Science, City University of Hong Kong, Hong Kong. From November 2020 to November 2021, he also worked as a Marie Curie Fellow (Granted by the Marie Skłodowska-Curie Individual Fellowships of European Commission) with the Faculty of Computing, Engineering and Media, De Montfort University, United Kingdom. From October 2021 to November 2021, he also worked as a visiting researcher (secondment of the Marie Skłodowska-Curie Individual Fellowships) with the Computer Vision and Graphics group, Fraunhofer Heinrich-Hertz-Institut (HHI), Germany. His current research interests include 3D visual coding, processing, and communication. He served as an Associate Editor for IEEE Transactions on Image Processing (since 2025), and Associate Editor for IEEE Transactions on Consumer Electronics (since 2024), an Associate Editor for IET Image Processing (since 2023), an Area Chair for IEEE ICME (since 2020).
\end{IEEEbiography}
\vspace{-5cm}
\begin{IEEEbiography}
[{\includegraphics[width=1in,height=1.25in,clip,keepaspectratio]{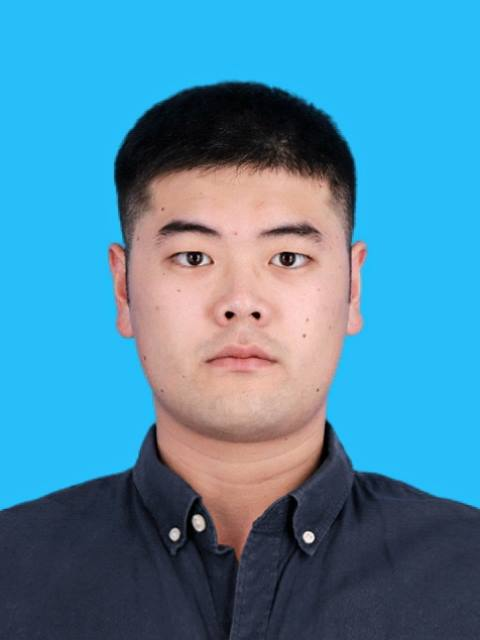}}] 
{Zhengxin Li}received the B.S. degree in automation from Shandong University, Weihai, China, in 2022. He is currently working toward the MS degree with the School of Control Science and Engineering, Shandong University, Jinan, China. His current research interests include video coding for machines, and point cloud coding for machines.
\end{IEEEbiography}
\vspace{-5cm}
\begin{IEEEbiography}
[{\includegraphics[width=1in,height=1.25in,clip,keepaspectratio]{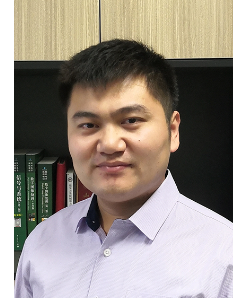}}] 
{Shuai Li} (SeniorMember, IEEE) received the Ph.D. degree from the University of Wollongong, Australia, in 2018. He was an Associate Professor with the School of Information and Communication Engineering, University of Electronic Science and Technology of China, China, from 2018 to 2020. He is currently a Professor and QiLu Young Scholar with the School of Control Science and Engineering, Shandong University (SDU), China. His research interests include image/video coding, 3D video processing, and computer vision. 
\end{IEEEbiography}
\vspace{-5cm}
\begin{IEEEbiography}
[{\includegraphics[width=1in,height=1.25in,clip,keepaspectratio]{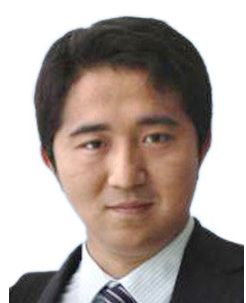}}] 
{Wei Zhang} (SeniorMember, IEEE) received the PhD degree in electronic engineering from the Chinese University of Hong Kong, in 2010. He is currently with the School of Control Science and Engineer ing, Shandong University, Jinan, China. His research interests include computer vision and robotics. He has served as a program committee member and a reviewer for various international conferences and journals 
\end{IEEEbiography}

\end{document}